\documentclass[conference]{IEEEtran}
\usepackage{multirow}
\IEEEoverridecommandlockouts
\usepackage{cite}
\usepackage{amsmath,amssymb,amsfonts,bm}
\usepackage{mathtools}
\usepackage{algorithmic}
\usepackage{optidef}
\usepackage{graphicx}
\usepackage{textcomp}
\usepackage{subcaption}
\usepackage{booktabs}
\usepackage{tabularx}
\usepackage{xcolor}
\usepackage{flushend}
\usepackage[normalem]{ulem}
\newcommand{\tightcell}[2]{%
  \begin{tabular}[t]{@{}c@{}}%
    #1\\[-3pt] {\tiny (#2)}%
  \end{tabular}%
}

\usepackage[ruled,vlined,linesnumbered]{algorithm2e}

\usepackage[hidelinks]{hyperref}
\hypersetup{
    colorlinks=true, 
    unicode=true, 
    linkcolor=blue!50!black, 
    citecolor=blue!50!black, 
    filecolor=blue!50!black, 
    urlcolor=blue!50!black
}

\def\BibTeX{{\rm B\kern-.05em{\sc i\kern-.025em b}\kern-.08em
    T\kern-.1667em\lower.7ex\hbox{E}\kern-.125emX}}
\begin{document}

\title{Green Federated Learning\\via Carbon-Aware Client and Time Slot Scheduling\vspace{-0.8cm}
\thanks{
This research was supported in part by the French government, through
the ``Plan de Relance" and the 3IA Côte d’Azur Investments in the Future project managed by the National Research Agency (ANR) with the reference number ANR-19-P3IA-0002, and in part by the European Network of Excellence dAIEDGE under Grant Agreement Nr. 101120726,
the EU HORIZON MSCA 2023 DN project FINALITY (G.A. 101168816), 
the Groupe La Poste, sponsor of the Inria Foundation, in the framework of the FedMalin Inria Challenge,
and the Wallenberg AI, Autonomous Systems and Software Program (WASP) funded by the Knut and Alice Wallenberg Foundation.
}
\thanks{Our code is available at: \url{https://github.com/chrdz/GreenFL}.}
}

\author{
\IEEEauthorblockN{Daniel Richards Arputharaj\IEEEauthorrefmark{1}, Charlotte Rodriguez\IEEEauthorrefmark{1}\IEEEauthorrefmark{2}, Angelo Rodio\IEEEauthorrefmark{3}, Giovanni Neglia\IEEEauthorrefmark{1}}
\IEEEauthorblockA{
\IEEEauthorrefmark{1}Inria Centre at Universit\'e C\^ote d’Azur, France. Email: \{firstname.lastname\}@inria.fr, \\
\IEEEauthorrefmark{2}Accenture, France. Email: {c.q.rodriguez@accenture.com}, \\
\IEEEauthorrefmark{3}Department of Electrical Engineering (ISY), Link\"oping University, Sweden. Email: {angelo.rodio@liu.se}}
}

\maketitle
\bstctlcite{BSTcontrol}
\thispagestyle{plain}
\pagestyle{plain}

\begin{abstract}
Training large-scale machine learning models incurs substantial carbon emissions. Federated Learning (FL), by distributing computation across geographically dispersed clients, offers a natural framework to leverage regional and temporal variations in Carbon Intensity (CI). This paper investigates how to reduce emissions in FL through carbon-aware client selection and training scheduling. 
We first quantify the emission savings of a carbon-aware scheduling policy that leverages slack time---permitting a modest extension of the training duration so that clients can defer local training rounds to lower-carbon periods.
We then examine the performance trade-offs of such scheduling which stem from  statistical heterogeneity among clients, selection bias in participation, and temporal correlation in model updates.
To leverage these trade-offs, we construct a carbon-aware scheduler that integrates slack time, $\alpha$-fair carbon allocation, and a global fine-tuning phase. Experiments on real-world CI data show that our scheduler outperforms slack-agnostic baselines, achieving higher model accuracy across a wide range of carbon budgets, with especially strong gains under tight carbon constraints.
\end{abstract}

\section{Introduction}

With the growing complexity of machine learning (ML) models, training requires access to large-scale, high-quality datasets and high-throughput computing infrastructure---often involving hundreds of terabytes of data and thousands of GPU-hours over days or even weeks of continuous execution~\cite{strubell2019}.

In terms of \emph{energy consumption}, training a baseline convolutional neural network such as ResNet-50 on the ImageNet dataset for a single-GPU, 90-epoch run consumes 55–90 kilowatt-hours (kWh) on the Inria Nef cluster, comparable to the energy needed for an electric car trip from Nice to Paris~\cite{Inria_NEF}. 
The resulting environmental cost depends not only on energy consumption but also on the Carbon Intensity (CI) of the electricity used, typically expressed in CO\textsubscript{2}-equivalents per kilowatt-hour (CO\textsubscript{2}e/kWh). This standardized metric aggregates the global warming impact of all greenhouse gases (and not just CO\textsubscript{2}). CI varies across time and geography, reflecting differences in local energy mix. 
As the current level of resource consumption becomes increasingly unsustainable, research priorities are shifting from exclusively maximizing model accuracy to jointly optimizing predictive performance and energy efficiency~\cite{budennyy2022, strubell2019}, giving rise to the principles of Green ICT and Green AI~\cite{hankel2016, schwartz2020}.

In terms of \emph{data production}, training data is inherently decentralized---generated at the user level and constrained by geographic and jurisdictional boundaries. Centralizing such data is often infeasible due to bandwidth limitations, latency, and energy overheads associated with large-scale data transfer. In addition, regulatory frameworks such as the GDPR in Europe and the CCPA in the U.S. impose data sovereignty, regulating cross-border data transfer and confining training data to their region of origin. Together, these constraints result in statistical heterogeneity and non-IID data distributions across computing nodes, as local datasets reflect region-specific demographic, behavioral, and cultural characteristics.

\emph{Federated Learning (FL)} emerges as an effective paradigm for training ML models across geographically distributed computing nodes---also referred to as clients~\cite{fed1, fedavg, fed3}. Unlike centralized learning, FL avoids the transfer of raw data by exchanging model updates.

In this work, we explore the idea that FL’s decentralized structure can also be leveraged to reduce the environmental impact of training by strategically allocating it to clients in regions and periods with lower CI.
Indeed, due to disparities in regional energy mixes, CI varies significantly across both time and location. This variability creates opportunities for \emph{carbon-aware scheduling}, in which training is shifted toward low-carbon regions and time windows. 
This paper addresses the following question: 
\emph{Given a pool of geographically distributed clients characterized by heterogeneous energy efficiency and carbon intensity, how should we schedule client training to minimize the environmental impact?}
We quantify the potential for substantial carbon savings by leveraging slack time---the flexibility to extend the training process beyond its minimum duration and defer computation to low-carbon periods.

In addition to highlighting the potential benefits, this paper identifies and addresses the following challenges inherent to carbon-aware scheduling:

1) Due to statistical heterogeneity, a carbon-greedy scheduling that tends to exclude high-emission clients will introduce statistical bias in the learned model. Our carbon-aware scheduler allocates a strictly positive share of the global carbon budget to each client through an $\alpha$\emph{-fair scheduling policy}.

2) Due to geographic variability in CI, even fair carbon-aware scheduling induces skewed client selection during training, leading to selection bias. We use an \emph{unbiased aggregation rule} that compensates for heterogeneous client selection, ensuring all clients contribute equally to the final model.

3) Due to temporal correlations in CI, carbon-aware scheduling suffers from last-iterate bias: early updates are rapidly forgotten, while late-round updates disproportionately influence the final model. To mitigate this imbalance, our scheduler introduces a \emph{fine-tuning phase} with full client selection and optimizes its placement under the carbon budget.

We evaluate our approach through extensive simulations using real-world CI traces from Electricity Maps~\cite{ElectricityMaps}. The experimental results quantify the benefits of slack time, $\alpha$-fair carbon allocation, and fine-tuning, demonstrating consistent improvements over slack-agnostic training, particularly under tight carbon budget constraints.

\section{Related Works}
\label{sec:related_works}
As concerns grow over the environmental impact of ML training, driven by the increasing scale of models, research has focused on making ML more sustainable. Most efforts predominantly target centralized settings, particularly single-tenant data centers. 
At the hardware level, specialized accelerators can drastically reduce energy consumption per operation---often by orders of magnitude compared to traditional CPUs and GPUs~\cite{Myu_2023, Pham_2024}.
At the platform level, carbon-aware schedulers migrate workloads to low-carbon-intensity periods~\cite{Acun_2023, Bashir_2021}. Orthogonal model-centric techniques, such as pruning, quantization, and distillation, reduce floating-point operations and memory usage with relatively minor impact on accuracy \cite{Gholami_2018, Sanh_2019, Sun_2024}.

While effective in centralized settings, these strategies are often ill-suited for federated learning, where training is distributed across statistically heterogeneous, geographically dispersed clients. A large body of work in FL optimizes client selection to reduce wall-clock time or iteration complexity.
\emph{System-oriented} approaches (e.g., \cite{nishioClientSelectionFederated2019, REFL, TiFL}) aim to accelerate convergence by selecting clients based on gradient norm \cite{Chen_2021} or loss-based utilities~\cite{Cho_2020}. 
\emph{Fairness-oriented} approaches (e.g., \cite{liFairResourceAllocation2019, Oort}), enforce diverse client selection by prioritizing clients with high statistical utility. However, all these works overlook the carbon footprint of the resulting schedules.

A parallel line of research has investigated how carbon-aware scheduling can reduce the environmental impact of federated learning. 
Early work shows that deferring training to low-carbon-intensity periods or selecting clients based on emission profiles can significantly reduce emissions \cite{qiu2023carbon, dodge}. 

Building on these foundations, recent works like FedZero \cite{wiesner2024fedzero}, FedCarbon \cite{fedcarbon}, CAFE \cite{Bian_2024}, and GREED \cite{albelaihiEnergyAwareFL} jointly optimize convergence and carbon efficiency by scheduling FL rounds during periods and in regions of low CI. 
Recognizing the bias introduced by statistical heterogeneity, FedZero and CAFE incorporate fairness into carbon-aware client selection by promoting diverse participation and prioritizing clients with high statistical utility.
However, estimating this utility requires all clients to compute local gradients on the current model at every round, regardless of whether they are selected or not---a common assumption in these works---which incurs an additional carbon cost. While this overhead is negligible in cross-silo settings (e.g., data centers), it becomes significant in cross-device scenarios, where clients are resource-constrained (e.g., smartphones) and both computation and communication are costly.
In contrast, FedCarbon and GREED avoid this overhead by omitting fairness criteria from client selection, but consequently overlook the learning bias that carbon-aware scheduling can introduce.

Unlike prior work, this paper incorporates fair client selection without relying on statistical utility, and is therefore applicable to both cross-silo and cross-device settings.
In the latter setting, where individual devices may not always be available, 
selecting clients from a region can be intended as defining a sampling probability over that region.
To our knowledge, we are the first to address \emph{temporal bias} in carbon-aware scheduling, a challenge stemming from correlations in CI profiles, yet overlooked in existing literature.
To our knowledge, we are the first to address \emph{temporal bias} in carbon-aware scheduling, a challenge stemming from correlations in CI profiles, yet overlooked in existing literature.

\section{Problem Description}
\label{sec:prob_desc}

A central server coordinates a set of clients $\mathcal{C} \coloneqq \{1, \dots, K\}$ to collaboratively learn the parameters $\theta \in \mathbb{R}^d$ of a global ML model (e.g., the weights of a neural network architecture). Each client $c \in \mathcal{C}$ holds a private local dataset $D_c$ and evaluates the quality of model parameters $\theta$ on data sample $z \in D_c$ via a loss function $\ell(\theta; z)$. The global learning objective is to minimize the average empirical loss across all clients:
\begin{align}
    F(\theta) \coloneqq \frac{1}{K} \sum_{c \in \mathcal{C}} \Biggl[ F_c(\theta) \coloneqq \frac{1}{|D_c|} \sum_{z \in D_c} \ell(\theta; z) \Biggr]. \label{eq:g_obj}
\end{align}

\subsection{The FedAvg Algorithm}

Problem~\eqref{eq:g_obj} is commonly solved using iterative algorithms such as \emph{Federated Averaging (FedAvg)}~\cite{fedavg}, which proceed over $T$ communication rounds between the server and clients.

At each round $t \in \{1,\dots,T\}$, the server selects a subset of clients $\mathcal{A}^{(t)} \subseteq \mathcal{C}$ and broadcasts the current global model~$\theta^{(t)}$. Let $a_c^{(t)} \in \{0, 1\}$ denote the binary indicator of whether client~$c$ is selected at round $t$, so that $a_c^{(t)} = 1$ if $c \in \mathcal{A}^{(t)}$, and~$0$ otherwise.
Each selected client $c \in \mathcal{A}^{(t)}$ performs $\tau$ local stochastic gradient descent (SGD) steps on its local dataset:
\begin{align}
    \theta_c^{(t,l+1)} = \theta_c^{(t,l)} - \eta \nabla F_c(\theta_c^{(t,l)}, \mathcal{B}_c^{(t,l)}),~l = 0, \dots, \tau - 1,
\end{align}
where \( \eta > 0 \) is the learning rate, \( \mathcal{B}_c^{(t,l)} \subseteq D_c \) is a randomly sampled mini-batch, and $\nabla F_c(\cdot, \mathcal{B}) \coloneqq \frac{1}{|\mathcal{B}|} \sum_{z \in B} \nabla \ell(\cdot, z)$ is an unbiased gradient estimate of \( \nabla F_c (\cdot) \).
Each client $c \in \mathcal{A}^{(t)}$ then returns its local update $\Delta_c^{(t)} = \theta^{(t)} - \theta_c^{(t,\tau)}$ to the server, which aggregates the updates as:
\begin{align}
    \Delta_{\text{FedAvg}}^{(t)} = \frac{1}{|\mathcal{A}^{(t)}|} \sum_{c \in \mathcal{A}^{(t)}} \Delta_c^{(t)}, \label{eq:g_agg}
\end{align}
and updates the global model as $\theta^{(t+1)} = \theta^{(t)} - \Delta_{\text{FedAvg}}^{(t)}$.
Because clients are geographically distributed, different selection sequences $\{\mathcal{A}^{(t)}\}_{t=1}^{T}$ yield different carbon emissions, depending on the temporal and regional variability of CI.

\subsection{Carbon Footprint of FL Training}

As we mentioned above, the CI is measured in kilograms of carbon dioxide equivalent (CO\textsubscript{2}e) per kWh of energy consumed (kgCO\textsubscript{2}e/kWh).
Platforms such as Electricity Maps~\cite{ElectricityMaps} provide CI estimates, including both hourly historical data and 72-hour forecasts, for a broad set of geographic regions.
In this paper, for the sake of concreteness, we consider historical CI data from 2022
across 54 geographic regions, as provided by~\cite{ElectricityMaps}.
For simplicity, we assume that each communication round corresponds to a one-hour time slot $t \in \{ 1, \dots, T \}$. If the round duration differs, the CI granularity should be adjusted accordingly.
Moreover, each client \( c \) is characterized by a fixed power draw \( P_c \) (in kW), leading to the same energy consumption over each time slot \( E_c = P_c \times 1 \) (in kWh). Thus, the carbon footprint \( g_c^{(t)} \) incurred by selecting client \( c \) during the interval \( [t, t+1] \) is approximated by $g_{c}^{(t)} = E_c \times \text{CI}_{c}^{(t)}$,
where \( \text{CI}_{c}^{(t)} \) is the average CI of client $c$ during time slot $t$.

Our objective is to design a carbon-aware FL schedule that, under a predefined carbon budget~$k$ (in kgCO\textsubscript{2}e), jointly decides \emph{which clients to select} and \emph{at which time slots}, while preserving model accuracy.
A key challenge here is that clients may have possibly diverse datasets, and the underlying energy sources are inherently heterogeneous and correlated across time and geography. These combined effects influence client selection strategies in a nontrivial way.

\begin{table}[t]
\centering
\footnotesize
\caption{List of Symbols}
\label{tab:symbols}
\begin{tabularx}{\columnwidth}{@{} l X @{}}
\toprule
\textbf{Symbol} & \textbf{Description} \\
\midrule
$K$ & Total number of clients \\
$\mathcal{C}$ & Set of clients, $\mathcal{C}\coloneqq\{1,\dots,K\}$ \\
$D_c$ & Local dataset of client $c$ \\
$\ell(\theta; z)$ & Loss of model $\theta$ on data sample $z$ \\
$F_c(\theta)$ & Local objective of client $c$, $F_c(\theta)\coloneqq\frac{1}{|D_c|}\sum_{z\in D_c}\ell(\theta;z)$ \\
$F(\theta)$ & Global objective, $F(\theta)\coloneqq\frac{1}{K}\sum_{c\in\mathcal{C}}F_c(\theta)$ \\
$T$ & Number of communication rounds \\
$\mathcal{A}^{(t)}$ & Subset of clients selected at round/slot $t$ \\
$a_c^{(t)}$ & Binary indicator: $1$ if client $c$ selected at slot $t$, else $0$ \\
$\tau$ & Number of local steps \\
$\eta$ & Learning rate \\
$\mathcal{B}c^{(t,l)}$ & Mini-batch sampled by client $c$ at round $t$, local step $l$ \\
$\nabla F_c(\cdot,\mathcal{B})$ & Unbiased mini-batch gradient estimator for client $c$ \\
$F^*$ & Minimum value of the global objective $F$\\
$\theta_c^{(t,l)}$ & Local model of client $c$ at round $t$, local step $l$ \\
$\Delta_c^{(t)}$ & Local update from client $c$ at round $t$, $\Delta_c^{(t)}=\theta^{(t)}-\theta_c^{(t,\tau)}$ \\
$\Delta_{\text{FedAvg}}^{(t)}$ & FedAvg update at round $t$ (Eq.~\eqref{eq:g_agg}) \\
$\Delta_{\text{U-FedAvg}}^{(t)}$ & Unbiased FedAvg update at round $t$ (Eq.~\eqref{eq:g_agg_unbiased}) \\
$P_c$ & Power draw of client $c$ \\
$E_c$ & Energy consumed per time slot by client $c$ \\
$\mathrm{CI}_c^{(t)}$ & Carbon intensity at client $c$ during slot $t$  \\
$g_c^{(t)}$ & Carbon cost of selecting $c$ at $t$, $g_c^{(t)}=E_c\cdot\mathrm{CI}_c^{(t)}$ \\
$g_{\text{max}}$ & Maximum carbon cost across all clients and time slots \\
$k$ & Global carbon budget \\
$t_{\mathrm{sl}}$ & Slack time (number of extra time slots beyond $T$) \\
$\mathcal{T}_c$ & Set of $T$ time slots for client $c$ within window of length $T+t_{\mathrm{sl}}$ \\
$\Delta_{\text{CO\textsubscript{2}e}}(c)$ & Relative CO\textsubscript{2}e savings for client $c$ (Eq.~\eqref{eq:saving_c}) \\
$\Delta_{\text{CO\textsubscript{2}e}}(N)$ & Relative CO\textsubscript{2}e savings for $N$ clients with slack vs without slack \\
$N$ & Number of clients selected (varies between $1$ and $K$) \\
$\mathcal{S}_N^{\text{fixed}}$ & Set of $N$ clients selected without using slack time \\
$\mathcal{S}_N^{\text{slack}}$ & Set of $N$ clients selected when slack is allowed \\
$\alpha$ & Fairness parameter in $\alpha$-fair carbon allocation, $\alpha \in (0,1]$\\
$\pi_c$ & Selection frequency of client c $\pi_c \coloneqq \frac{1}{T} \sum_{t=1}^{T} a_{c}^{(t)}$\\
$\bm{\pi}$ & Vector of selection frequencies $\bm{\pi}\coloneqq(\pi_1,\dots,\pi_K)$\\
$\rho_{\text{H}}$ & Selection heterogeneity measure, $\rho_{\text{H}} \coloneqq \frac{1}{K} \sum_{c=1}^{K} \frac{1-\pi_c}{\pi_c}$ \\
$\rho_{\text{T}}$ & Temporal correlation measure in client selection \\
$\rho_{\text{TS}}$ & Temporal-spatial correlation measure (second-largest eigenvalue of transition matrix) \\
$t_{\mathrm{ft}}$ & Fine-tuning duration (number of rounds) \\
$s$ & Fine-tuning end time \\
$\mathcal{F}(s)$ & Fine-tuning window, $\mathcal{F}(s) = \{T+s-t_{\mathrm{ft}}+1, \dots, T+s\}$\\
$A$ & Scheduling matrix $A \coloneqq (a_{c}^{(t)})_{c,t}$\\
\bottomrule
\end{tabularx}
\end{table}

\section{Prospective Carbon Savings via Slack Time}
\label{sec:prelims}

\emph{How much CO\textsubscript{2}e can FL potentially avoid by deferring training to periods of low carbon intensity?}

\begin{figure}[t]
\centering
  \includegraphics[width=0.89\columnwidth]{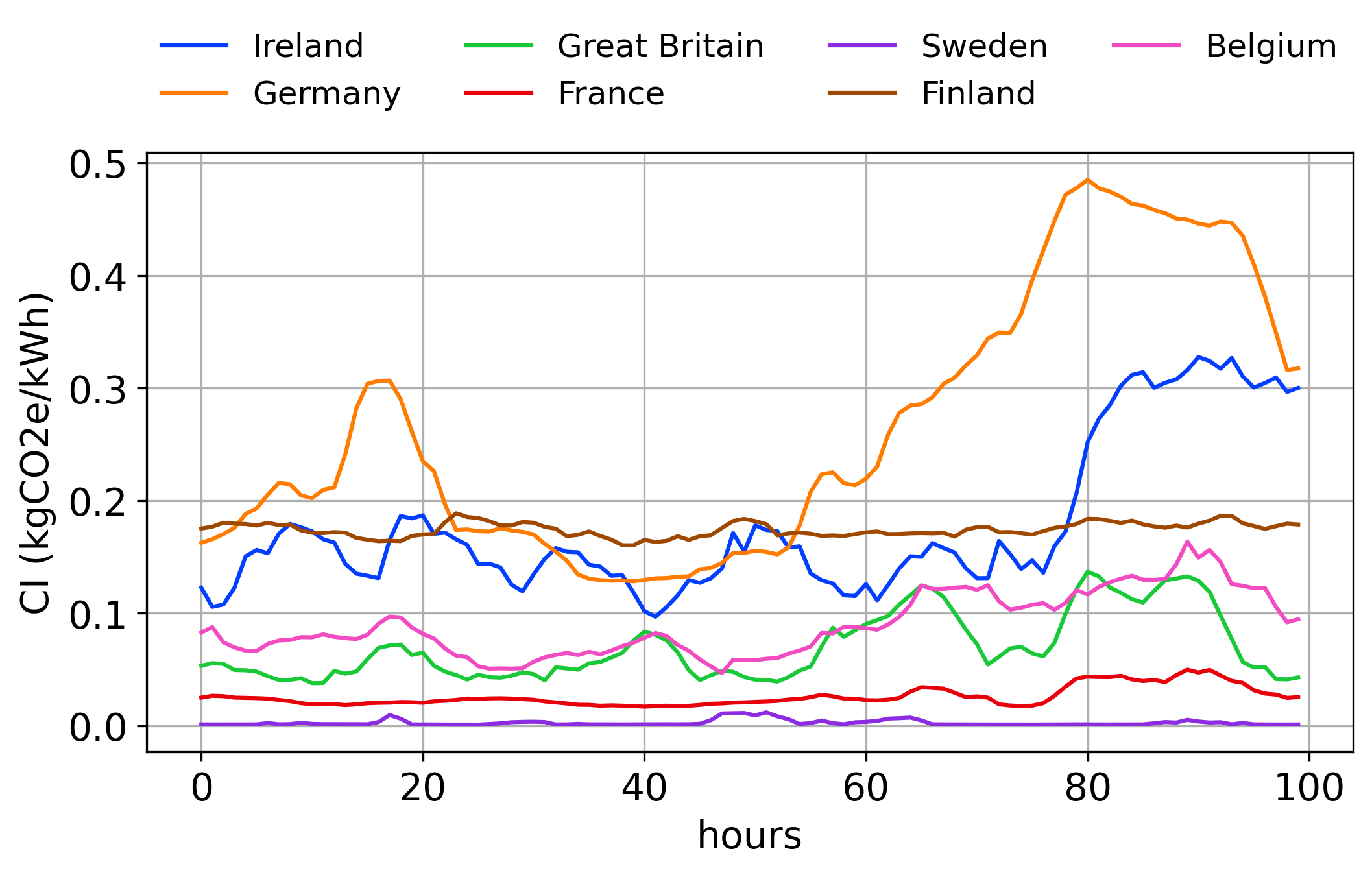}
  \caption{CI time evolution across selected countries from January 1 to January 5, 2022.
  }
  \label{fig:raw_CI_data}
\end{figure}

Figure~\ref{fig:raw_CI_data}, sourced from the Electricity Maps~\cite{ElectricityMaps}, shows the hourly variation in CI across selected European countries over five consecutive days in January 2022.
Countries like France and Sweden exhibit low, stable CI levels below 0.05 kgCO\textsubscript{2}e/kWh, while countries like Germany and Ireland show the highest variability with fluctuations above 0.2 kgCO\textsubscript{2}e/kWh. This variability in the CI profiles across countries opens prospective opportunities for emissions reduction, as training can be partially offloaded to low-CI periods.

We formalize these potential savings by defining \emph{slack time}, $t_\mathrm{sl}$, as the number of extra time slots available beyond the strictly necessary training duration $T$. A larger slack time offers more flexibility to defer training to low-CI periods, thereby reducing carbon emissions.

In what follows, we quantify the potential benefits of strategically scheduling client training, while temporarily ignoring its impact on final model quality.

\subsection{Impact of Slack Time on Individual Clients}
\label{sec:single_client_slack_time}

\begin{figure}[t]
\centering
  \includegraphics[width=\columnwidth]{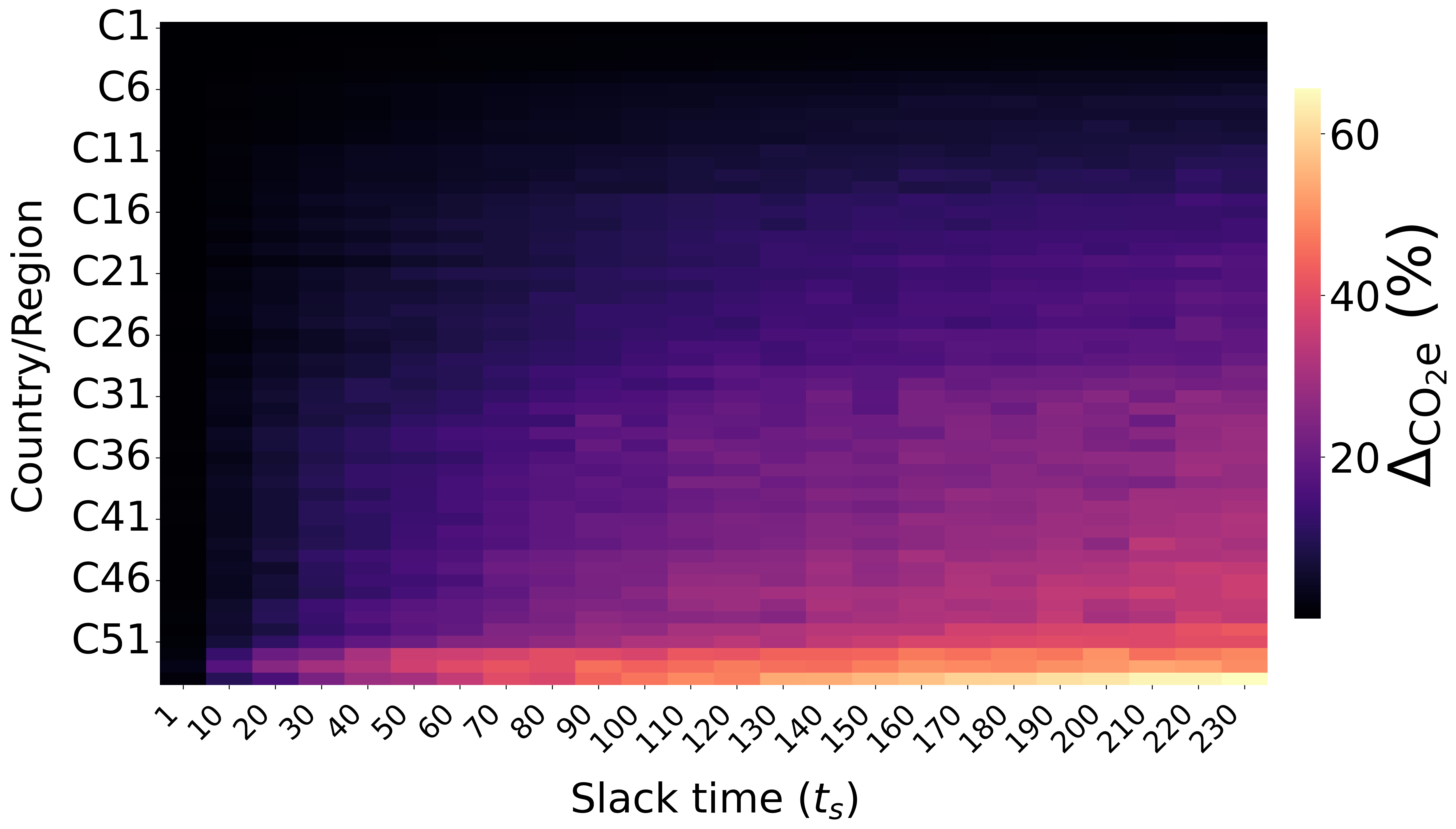}
  \caption{CO\textsubscript{2}e savings per individual client as a function of slack time. Each row corresponds to a different region.}
  \label{fig:slack_heat_map}
\end{figure}

We first quantify CO\textsubscript{2}e savings for an \emph{individual client.}
We set a fixed training duration of $T=100$ hours and progressively increase slack time $t_\mathrm{sl} \in [1, 236]$, yielding a scheduling window of up to two weeks ($T + t_\mathrm{sl} = 336$ hours). For each client $c$, we select a set $\mathcal{T}_{c}$ consisting of the $T$ time slots with the lowest CI within the given window, i.e., $\mathcal{T}_{c} \in \arg\min_{\mathcal{T}' \subseteq \{1, \dots, T + t_\mathrm{sl}\},|\mathcal{T}'| = T} \sum_{t \in \mathcal{T}'} g_c^{(t)}$. We measure the CO\textsubscript{2}e savings relative to the first $T$ slots (i.e., without slack time) as: 
\begin{align}
    \Delta_{\text{CO\textsubscript{2}e}}(c)  = \left(1 - \frac{ \sum_{t \in \mathcal{T}_c} g_{c}^{(t)}}{ \sum_{t'=1}^T g_{c}^{(t')} }\right). \label{eq:saving_c}
\end{align}

Figure~\ref{fig:slack_heat_map} shows that clients with high carbon intensity variability can already achieve a 20\% reduction in emissions when training is allowed to extend by just $t_\mathrm{sl} = 20$ extra hours, and up to 60\% reductions with $t_\mathrm{sl} = 236$ hours.
Moreover, with $t_\mathrm{sl} = 236$ hours, approximately 80\% of clients reduce their training carbon intensity by at least 10\%, and 50\% of them achieve reductions by at least 20\%.

\subsection{Impact of Slack Time and Client Selection}
\label{sec:mult_client_slack_time}

\begin{figure}[t]
\centering
  \includegraphics[width=\columnwidth]{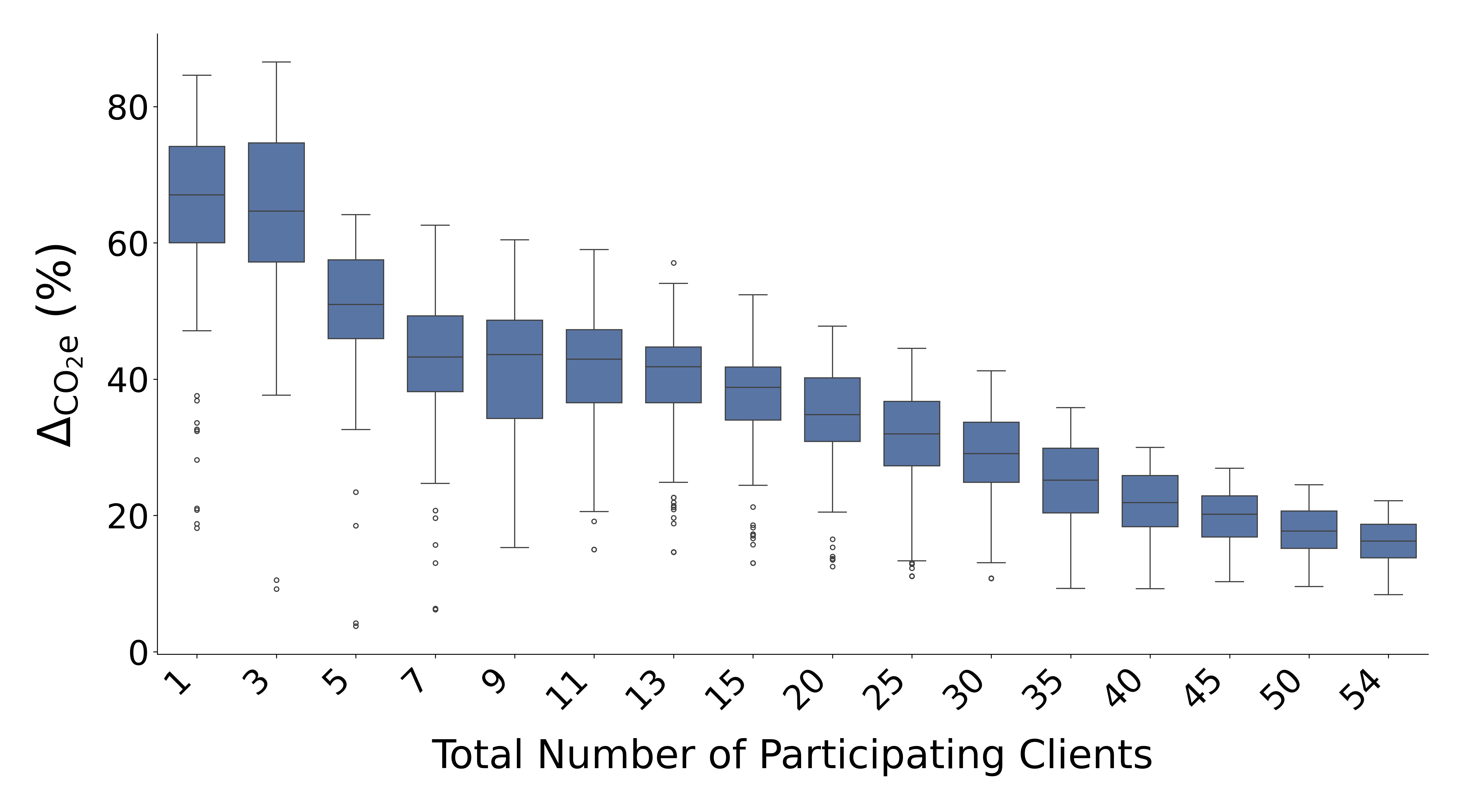}
  \caption{CO\textsubscript{2}e savings from slack-aware versus slack-agnostic client selection as a function of the number of selected clients (with lowest average carbon emissions).}
  \label{fig:delta_analysis}
\end{figure}

We extend the analysis to jointly optimize \emph{client selection} and \emph{time slot scheduling}.
We again fix $T = 100$ hours and consider the maximal slack time $t_\mathrm{sl} = 236$ hours, varying the number of selected clients $N$ from 1 to $K=54$. 
We compare two client selection strategies: 
\begin{itemize}
    \item \emph{Without slack time}, where we select the set of $N$ clients $\mathcal{S}_N^{\text{fixed}}$ with the lowest cumulative carbon emissions over the initial $T$-hour training period. Formally, $\mathcal{S}_N^{\text{fixed}} \in \arg\min_{\mathcal{S} \subseteq \{1,\dots,K\},\ |\mathcal{S}| = N} \sum_{c \in \mathcal{S}} \sum_{t=1}^{T} g_c^{(t)}$.
    \item \emph{With slack time}, where we select the set of clients $\mathcal{S}_N^{\text{slack}}$ with the lowest emissions over their $T$ least carbon-intensive time slots within the extended scheduling window of length $T + t_\mathrm{sl}$. Formally, $\mathcal{S}_N^{\text{slack}} \in \arg\min_{\mathcal{S} \subseteq \{1,\dots,K\},\ |\mathcal{S}| = N} \sum_{c \in \mathcal{S}} \sum_{t \in \mathcal{T}_c} g_c^{(t)}$.
\end{itemize}
Figure~\ref{fig:delta_analysis} shows the relative reduction in \text{CO\textsubscript{2}e}:
\begin{align}
    \Delta_{\text{CO\textsubscript{2}e}}(N) = \left( 1- \frac{\sum_{c\in\mathcal{S}_N^{\text{slack}}} \sum_{t\in\mathcal{T}_c} g_{c}^{(t)}}{\sum_{c' \in \mathcal{S}_N^{\text{fixed}}} \sum_{t'=1}^{T} g_{c'}^{(t')}} \right). 
\end{align}

To account for seasonality effects, we averaged results over 100 randomly sampled training start times throughout the year 2022. As $N$ grows, emissions necessarily increase because each additional client contributes an equal or greater carbon cost. Nonetheless, accounting for slack time allows for meaningful reductions in overall carbon usage. CO\textsubscript{2}e savings exceed 60\% with fewer than 5 clients, remain above 40\% with up to 15 clients, and stabilize around 20\% when all $K=54$ clients are selected.

\section{Performance Trade-Offs in Carbon-Aware Scheduling}
\label{sec:exp_w_wo_ft}

The slack analysis showed substantial carbon savings assuming equal contribution of \emph{all} training slots to model performance.
In practice, the contribution of each slot depends on \emph{when} it occurs and on \emph{which} client performs it.
This section quantifies the practical trade-offs between carbon savings and model accuracy,
under a consistent experimental setup.

\medskip

\noindent\emph{Experimental Setup.} We simulate a FL environment with $K=7$ clients, each mapped to a different geographic region (Belgium, Great Britain, Ireland, Finland, Sweden, Germany, and France). Unless otherwise mentioned, training begins on January~1, 2022 and proceeds for $T = 50$ communication rounds, and each region is assigned hourly carbon intensity from real-world Electricity Maps~\cite{ElectricityMaps}. 
Each client trains a convolutional neural network (CNN) with $\sim\!\!\!1.2$ million parameters for MNIST classification, consisting of two $3\times 3$ convolutional layers with 32 and 64 filters (each followed by ReLU and 
$2\times 2$ max-pooling) and two fully connected layers with 128 hidden units and 10 output units. To simulate statistical heterogeneity, we partition the data across clients in a non-IID fashion using a Dirichlet distribution with concentration parameter $\beta = 0.5$. Clients perform $\tau = 5$ local SGD updates per round using mini-batches of size 128. Optimization uses cross-entropy loss with learning rates tuned for each experiment via grid search over $\eta \in \{ 10^{-1}, 10^{-1.5}, 10^{-2} \}$. All results are averaged over three independent runs.

\subsection{Fair Carbon Allocation under Statistical Heterogeneity}
\label{ssec:alpha_fair}

\begin{figure}[t]
  \centering
  \begin{subfigure}[t]{0.48\linewidth}
    \centering
    \includegraphics[height=2.8cm]{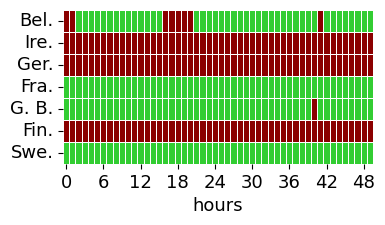}
    \caption*{\hspace{3.2em} (a) $\alpha = 1.0$}
  \end{subfigure}
  \hfill
  \begin{subfigure}[t]{0.48\linewidth}
    \centering
    \includegraphics[height=2.8cm]{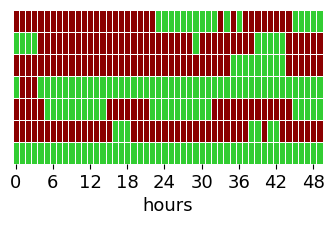}
    \caption*{(b) $\alpha = 10^{-3}$}
  \end{subfigure}
  \caption{Effect of fairness parameter $\alpha$ on allocation (green: selected, red: excluded): from a carbon-greedy (Fig.~\ref{fig:avmat0.001-1.0}a) to a carbon-fair (Fig.~\ref{fig:avmat0.001-1.0}b) allocation.}
  \label{fig:avmat0.001-1.0}
\end{figure}

In real-world FL systems, time slots are not interchangeable across clients due to statistical heterogeneity in their local datasets $\{D_c\}_{c=1}^K$, which differ in both sample size and distribution.
For instance, clients in different geographic regions may exhibit distinct data influenced by cultural, linguistic, or behavioral factors.
As a result, the local losses $\{F_c(\theta) \coloneqq \frac{1}{|D_c|} \sum_{z\in D_c} \ell(\theta; z)\}_{c=1}^K$ can deviate significantly from the global loss $F(\theta)\coloneqq\frac{1}{K}\sum_{c=1}^{K}F_c(\theta)$. 

A carbon-aware scheduler that selects client $c$ at time $t$ ($a_{c}^{(t)}=1$) minimizes, over $T$ rounds, the empirical loss $\tilde{F}_{T}(\theta) \coloneqq \frac{1}{KT} \sum_{c=1}^{K} \sum_{t=1}^{T} a_{c}^{(t)} F_c(\theta)$. If a high-emission client $c'$ is systematically excluded (i.e., $a_{c'}^{(t)} = 0$ for all $t$), its local loss $F_{c'}(\theta)$ does not contribute to $\tilde{F}_T(\theta)$, inducing a \emph{statistical bias} in the learned model.

More generally, under a fixed carbon budget~$k$, selecting only the lowest CI time slots may result in a skewed representation of clients, potentially harming model generalization.
To mitigate this statistical bias, each client should be guaranteed a non-zero fraction of the global carbon budget~$k$, while still favoring lower-emission clients. We formalize this trade-off using an $\alpha$-fair utility function~\cite{jain1984quantitative},
and define the resulting policy as the $\alpha$\emph{-fair carbon-aware scheduler}:
\begin{maxi!}
    {A \coloneqq (a_{c}^{(t)})_{c,t}}{\sum_{c=1}^{K} \biggl( \sum_{t=1}^{T + t_\mathrm{sl}} (g_{\max} - g_{c}^{(t)}) a_{c}^{(t)} \biggr)^{\alpha}}
    {\label{opt:alpha}}{}
    \addConstraint{\sum_{c=1}^{K} \sum_{t=1}^{T + t_\mathrm{sl}} g_{c}^{(t)} a_{c}^{(t)}}{\leq k}
    \addConstraint{a_{c}^{(t)}}{\in \{0,1\}, \quad \forall c, t},
\end{maxi!}
where $g_{\max} \coloneqq \max_{c,t} g_{c}^{(t)}$ is a constant that allows us to formulate the problem as a maximization task---the rationale for preferring a maximization formulation over minimization will be discussed in Section~\ref{sec:algo}.
The parameter $\alpha \in (0,1]$ controls the trade-off between pure carbon-efficiency and fair carbon allocation across clients: when \mbox{$\alpha = 1$}, the solution prioritizes selecting as many low-emission time slots as the budget allows, whereas as \mbox{$\alpha \to 0^+$}, the allocation tends toward distributing the carbon budget as equally as possible among clients at the cost of selecting a smaller number of time slots. Throughout the remainder of this work, we refer to the $\alpha = 1$ allocation as \emph{carbon-greedy}, and to allocations with smaller $\alpha$ as (more) \emph{carbon-fair}.

Figure~\ref{fig:avmat0.001-1.0} shows the resulting matrix $A \coloneqq (a_{c}^{(t)})_{c,t}$ for $\alpha = 1$ and $\alpha = 10^{-3}$, respectively. At $\alpha = 1$, 
high-emission clients are systematically excluded. In contrast, at $\alpha = 10^{-3}$, these clients are occasionally selected, at the cost of a 14\% reduction in the total number of scheduled training slots.

\begin{figure}[t]
  \centering
  \includegraphics[width=\columnwidth]{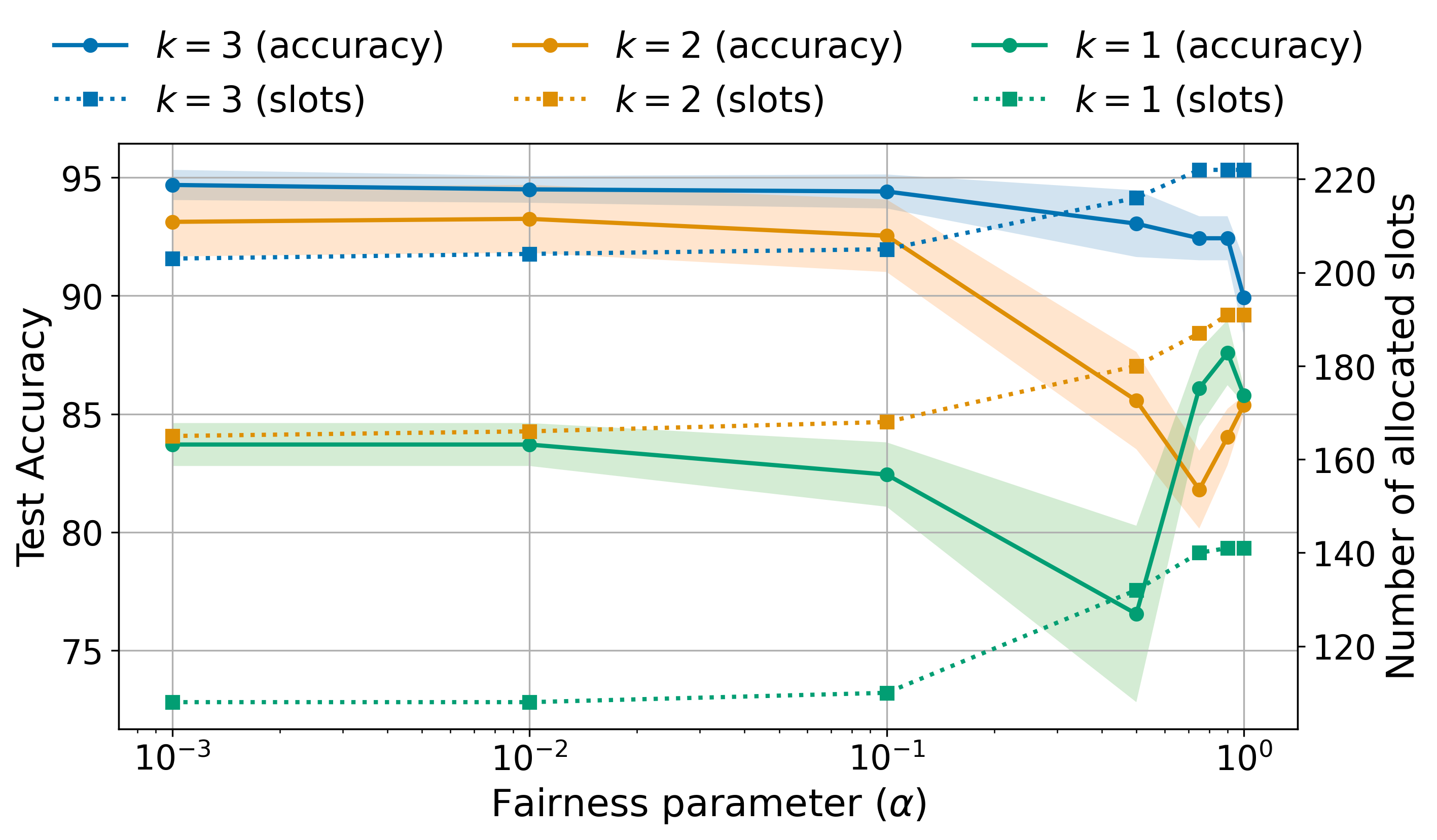}
  \caption{Effect of fairness parameter $\alpha$ on test accuracy for different carbon budgets: from a carbon-fair ($\alpha=10^{-3}$) to a carbon-greedy ($\alpha=1$) per client allocation.}
  \label{fig:alphaF_var}
\end{figure}

Figure~\ref{fig:alphaF_var} shows the trade-off between test accuracy and the fairness parameter~$\alpha$ under varying carbon budgets. 
Overall, we observe that for large (3 kgCO\textsubscript{2}e) and medium (2 kgCO\textsubscript{2}e) carbon budgets, increasing fairness (i.e., using smaller $\alpha$ values) reduces statistical bias and leads to substantial improvements in accuracy compared to the carbon-greedy case ($\alpha = 1$): 
+5 percentage points (pp) for large budgets and +8 pp for medium budgets. However, under tight carbon budgets (e.g., below 1~kgCO\textsubscript{2}e), enforcing fairness can limit the number of available training slots, ultimately leading to a drop in accuracy. 
Interestingly, within the range $\alpha \in [10^{-3}, 10^{-1}]$, performance remains relatively stable, indicating that the method is not overly sensitive to the exact choice of $\alpha$. This suggests that selecting a suitable value is not particularly challenging in practice—though it may come at the cost of suboptimal performance under very tight carbon budgets.

\subsection{Unbiased FedAvg under Client Selection Heterogeneity}

While the $\alpha$-fair scheduler in Opt.~\eqref{opt:alpha} guarantees each client a non-zero share of the carbon budget, it does not enforce uniform selection frequencies. Even as $\alpha \to 0^+$, the scheduler tends to equalize the carbon budget allocated per client, but clients with higher CI will be selected less frequently, as each of their training rounds consumes more of the budget. As a result, \emph{selection heterogeneity} can emerge: lower-emission clients can be selected significantly more frequently than high-emission ones, even at small $\alpha$.
For example, in Figure~\ref{fig:avmat0.001-1.0}b, low-carbon clients are selected up to 6 times more often than high-carbon ones when $\alpha=10^{-3}$. 

We define selection frequencies $\bm{\pi}\coloneqq(\pi_1,\dots,\pi_K)$, where $\pi_c \coloneqq \frac{1}{T} \sum_{t=1}^{T} a_{c}^{(t)}$ is the fraction of training rounds in which client $c$ is selected.
We measure selection heterogeneity as \mbox{$\rho_{\text{H}} \coloneqq \frac{1}{K} \sum_{c=1}^{K} \frac{1-\pi_c}{\pi_c}$}, with higher $\rho_{\text{H}}$ indicating more heterogeneous client selection. In practice, substantial disparities may persist even under carbon-fair scheduling: in Figure~\ref{fig:avmat0.001-1.0}b, we measure $\rho_{\text{H}}=6$. 

Selection heterogeneity can significantly degrade model quality. The standard FedAvg aggregation rule, \mbox{$\Delta_{\text{FedAvg}}^{(t)} = \frac{1}{|\mathcal{A}^{(t)}|} \sum_{c \in \mathcal{A}^{(t)}} \Delta_c^{(t)}$}, effectively minimizes the weighted loss $\tilde{F}_{T}(\theta) \coloneqq \frac{1}{K} \sum_{c=1}^{K} \pi_c F_c(\theta)$, where clients selected more frequently---those with lower CI---receive higher weight (this argument can be formalized by assuming that each client~$c$ is selected at each time slot with probability~$\pi_c$). The  selection bias introduces then a mismatch between the empirical loss $\tilde{F}_{T}(\theta)$ and the true global loss $F(\theta)$, leading to a non-vanishing convergence error proportional to the total variation distance $\text{d}_{\text{TV}}(\bm{1}/K, \bm{\pi}/\lVert \bm{\pi} \rVert_1) = \frac{1}{2} \sum_{c=1}^{K} \lvert \frac{1}{K} - \frac{\pi_c}{\lVert \bm{\pi} \rVert_1} \rvert$~\cite{rodioFederatedLearningHeterogeneous2023}.

To correct for selection bias, a carbon-aware scheduler must adopt an \emph{unbiased aggregation rule}, in which each client's update is reweighted by the inverse of its selection frequency~$\pi_c$:
\begin{align}
    \Delta_{\text{U-FedAvg}}^{(t)} = \frac{1}{K} \sum_{c \in \mathcal{A}^{(t)}} \frac{\Delta_c^{(t)}}{\pi_c}. \label{eq:g_agg_unbiased}
\end{align}
This variant is known as \emph{Unbiased FedAvg (U-FedAvg)}~\cite{rodioFederatedLearningHeterogeneous2023} and leads to an unbiased model.

For smooth and strongly-convex objective and diminishing learning rates, U-FedAvg achieves $\mathbb{E}[F(\theta^{(T)})] - F^\star \leq \mathcal{O}(\rho_{\text{H}}/T)$, where $F^*$ denotes the minimum value of $F$, eliminating selection bias but converging more slowly when selection heterogeneity $\rho_{\text{H}}$ is large~\cite{rodioFederatedLearningHeterogeneous2023}.

\subsection{Fine-Tuning under Temporal and Spatial Correlation}

\begin{figure}[t]
  \centering
  \includegraphics[width=0.9\columnwidth]{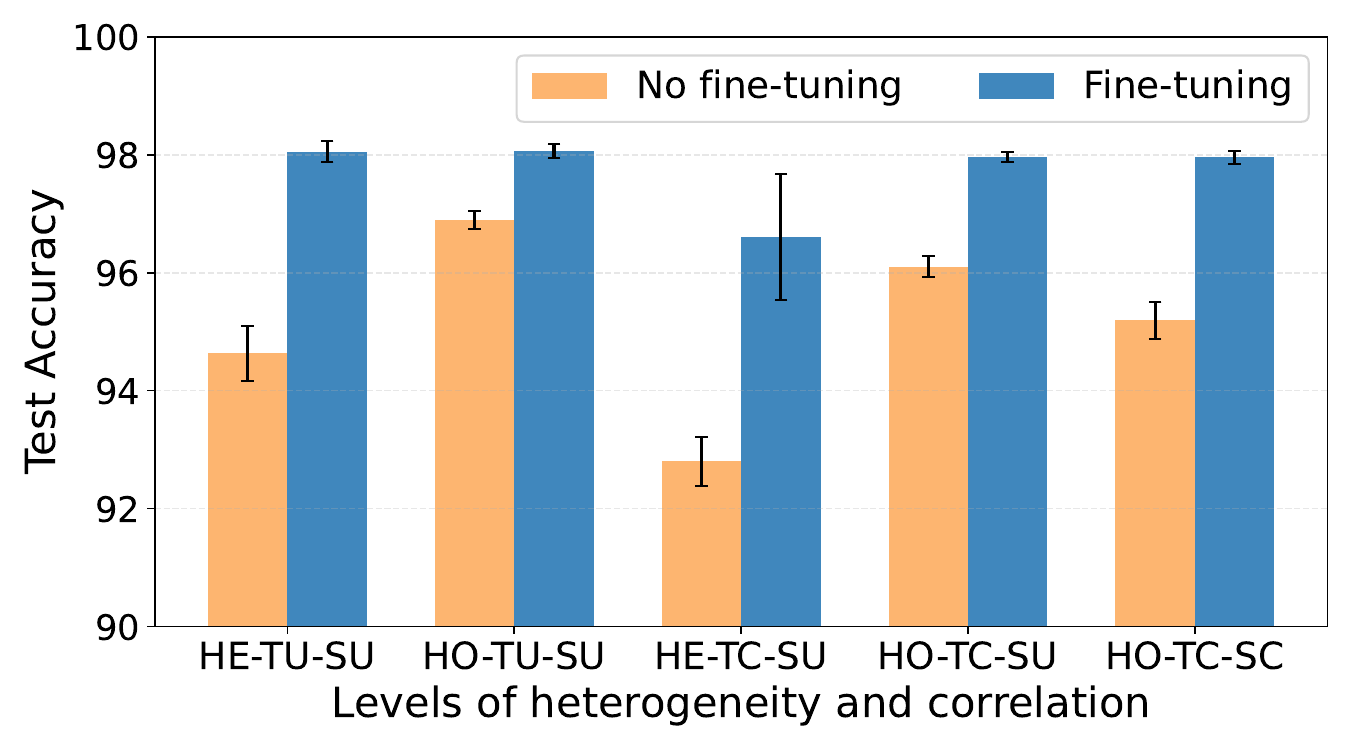}
  \caption{Effect of heterogeneity and correlation on test accuracy, with and without fine-tuning ($t_{\mathrm{ft}}=10$). Client selection is either homogeneous (HO: $\rho_{\text{H}}<1$) or heterogeneous (HE: $\rho_{\text{H}}>1.5$); time uncorrelated (TU: $\rho_{\text{T}}<0.1$) or time correlated (TC: $\rho_{\text{T}}>0.7$); and space uncorrelated (SU: $\rho_{\text{TS}}<0.35$) or space correlated (SC: $\rho_{\text{TS}}>0.45$).}
  \label{fig:ft_vs_noft}
\end{figure}

Temporal and spatial correlations in carbon intensity introduce additional challenges for carbon-aware scheduling. These correlations stem from systematic variations in regional energy mixes and demand.
For instance, in solar-heavy regions, CI drops predictably around midday when solar energy production peaks and lowers demand for fossil fuels.
In wind-dominated regions, CI fluctuates with weather conditions, while in coal-reliant grids, CI remains consistently high throughout the day.

The carbon-aware scheduler in Opt.~\eqref{opt:alpha} naturally inherits these correlated dynamics. \emph{Temporal correlation} arises when clients are consistently selected or excluded over long consecutive time periods, due to persistently low or high CI during specific hours or days. \emph{Spatial correlation} emerges when geographically close clients share similar CI patterns and are thus selected or excluded together. 
Correlated selection patterns are clearly visible in Figure~\ref{fig:avmat0.001-1.0}b: despite the strong fairness constraint imposed by $\alpha = 10^{-3}$, many clients tend to be either consistently selected or consistently excluded across long sequences of time slots.

These correlations degrade model performance by introducing temporal imbalance in the aggregation of client updates from different training rounds. 
Specifically, updates from clients selected in early rounds may be rapidly overwritten---a phenomenon known as \emph{catastrophic forgetting}~\cite{kirkpatrickOvercomingCatastrophicForgetting2017}---while updates from clients selected in later rounds typically exert a disproportionately high influence on the final model parameters, an effect known as \emph{last-iterate bias}~\cite{varreLastIterateConvergence2021}.

\begin{algorithm}[t]
\caption{Carbon-Aware FL Algorithm}
\label{alg:general_algo}

\textbf{Input: }{$\theta^{(1)}$, $\{D_c\}_{c=1}^{K}$, $\tau$, $\eta$, $(g_{c}^{(t)})_{c,t}$, $T$, $t_\mathrm{sl}$, $t_{\mathrm{ft}}$, $\alpha$ }

$((a_{c}^{(t)})_{c,t},s) \gets$ \emph{CarbonScheduler}($(g_{c}^{(t)})_{c,t}$, $t_\mathrm{sl}$, $t_{\mathrm{ft}}$, $\alpha$) \label{alg-line:scheduler}

\For{$t = 1, \dots, T + s$}{

    \For{$c = 1,\dots,K : a_{c}^{(t)} = 1$, in parallel\label{alg-line:client_opt}}{

    $\theta_c^{(t,0)} \gets \theta^{(t)}$

    \For{$l = 0,1\dots,\tau-1$}{
    
    $\theta_c^{(t,l+1)} = \theta_c^{(t,l)} - \eta \nabla F_c(\theta_c^{(t,l)}, \mathcal{B}_c^{(t,l)})$

    }

    $\Delta_{c}^{(t)} \gets (\theta^{(t)} - \theta_{c}^{(t,\tau)})$ \label{alg-line:client_update}

    }

    \If{$t \in \mathcal{F}(s)$}{
    
    $\Delta_{\text{\emph{FedAvg}}}^{(t)} = \frac{1}{K} \sum_{c=1}^{K} \Delta_c^{(t)}$ \label{alg-line:agg_fedavg}

    $\theta^{(t+1)} = \theta^{(t)} - \Delta_{\text{\emph{FedAvg}}}^{(t)}$ \label{alg-line:server_opt_fedavg}
    
    }
    
    \Else{

    $\Delta_{\text{\emph{U-FedAvg}}}^{(t)} = \frac{1}{K} \sum_{c=1}^{K} a_{c}^{(t)} \frac{\Delta_c^{(t)}}{\pi_c}$ \label{alg-line:agg_ufedavg}

    $\theta^{(t+1)} = \theta^{(t)} - \Delta_{\text{\emph{U-FedAvg}}}^{(t)}$ \label{alg-line:server_opt_ufedavg}
    
    }

}
\end{algorithm}

When client participation follows a Markov chain (MC) with state space $\{0,1\}^K$ and transition matrix $P$, \cite{rodioFederatedLearningHeterogeneous2023} shows that, for smooth and strongly convex objectives, U-FedAvg achieves the convergence bound $\mathbb{E}[F(\theta^{(T)})] - F^* \leq 
\mathcal O([\ln(1/\rho_{\text{TS}})T]^{-1})$, where $\rho_{\text{TS}}$ denotes the second-largest eigenvalue in modulus of $P$. This result highlights the impact of  correlations in client selection on the convergence rate, with slower-mixing chains (i.e., $\rho_{\text{TS}} \to 1$) leading to slower convergence.

Figure~\ref{fig:ft_vs_noft} reports test accuracy under varying levels of temporal and spatial correlation. In general, the MC can capture these correlations either separately or jointly. We denote $\rho_{\text{TS}}$ simply as $\rho_{\text{T}}$ when the MC exhibits only temporal correlation.
Compared to low-correlation settings ($\rho_{\text{T}} < 0.1$), test accuracy decreases by 1 pp when temporal correlation is high ($\rho_{\text{T}} > 0.7$), and by an additional 1 pp when spatial correlation is also high ($\rho_{\text{TS}}> 0.45$).

To compensate for last-iterate bias, we append a \emph{fine-tuning} window at the end of the training schedule. This window, defined as $\mathcal{F} = \{ T - t_{\mathrm{ft}}+ 1, \dots, T \}$, spans the final $t_{\mathrm{ft}}$ rounds and enforces full client selection ($a_{c}^{(t)}=1$ for all $c \in \mathcal{C}$, $t \in \mathcal{F}$). By imposing a final model update from every client, this final stage aims at mitigating the temporal imbalance caused by correlated selection and the dominance of late-round updates.
In Figure~\ref{fig:ft_vs_noft}, fine-tuning yields substantial gains, especially when both correlations are high ($\rho_{\text{TS}} > 0.45$), improving test accuracy by +3.8 pp.

\begin{table*}[t]

    \centering
    \caption{Final test accuracy under different carbon budget constraints ($k$) and fine-tuning end times ($s$). The first column reports the absolute available budget as well as percentage of the full-budget reference. The second and third columns show the training time ($T$) and the accuracy for the slack-agnostic baseline, respectively. The remaining columns correspond to varying fine-tuning end times $s$, evaluated for two fine-tuning durations ($t_{\mathrm{ft}}= 1$ and $t_{\mathrm{ft}}= 3$). Numbers in parentheses denote the standard deviation across three seeds. Bold values indicate the best accuracy in each row. A horizontal rule (---) indicates that placing fine-tuning at the given $s$ would exceed the available carbon budget and was therefore infeasible.}
    \label{tab:combined_table}
    \footnotesize
    \renewcommand{\arraystretch}{1.05} 
    \setlength{\tabcolsep}{1pt} 
        \begin{subtable}{\linewidth}
        \centering
        \caption{Test accuracy with fine-tuning duration $t_{\mathrm{ft}}=1$.}
        \label{tab:carbon_budget-1ft}
        \renewcommand{\arraystretch}{1.2}
        \setlength{\tabcolsep}{4pt}
        \begin{tabular}{|c|c|c|ccccccccccccccc|}
        \hline
        \multirow{2}{*}{\shortstack{\textbf{Carbon} \\ \textbf{Budget ($k$)}}} & \multirow{2}{*}{\shortstack{\textbf{Training} \\ \textbf{Time ($T$)}}} & \multirow{2}{*}{\shortstack{\textbf{No-Slack} \\ \textbf{Baseline}}} & \multicolumn{15}{c|}{\textbf{End time for fine-tuning ($s$)}} \\
        \cline{4-18} 
        & & & \textbf{1} & \textbf{3} & \textbf{5} & \textbf{7} & \textbf{9} & \textbf{10} & \textbf{20} & \textbf{30} & \textbf{40} & \textbf{50} & \textbf{70} & \textbf{90} & \textbf{110} & \textbf{130} & \textbf{150} \\
        \hline
7.98 / 82.0\% & 40 & \tightcell{98.82}{0.06} & \tightcell{98.86}{0.04} & \tightcell{98.82}{0.07} & \tightcell{98.86}{0.09} & \tightcell{98.83}{0.06} & \tightcell{98.86}{0.07} & \tightcell{98.91}{0.02} & \tightcell{98.95}{0.07} & \tightcell{98.91}{0.08} & \tightcell{98.99}{0.03} & \tightcell{99.01}{0.02} & \tightcell{98.97}{0.03} & \tightcell{98.97}{0.03} & \tightcell{\textbf{99.02}}{0.03} & \tightcell{98.98}{0.03} & \tightcell{98.98}{0.06} \\

6.2 / 63.67\% & 30 & \tightcell{98.66}{0.07} & \tightcell{98.68}{0.03} & \tightcell{98.72}{0.06} & \tightcell{98.71}{0.05} & \tightcell{98.78}{0.08} & \tightcell{98.8}{0.06} & \tightcell{98.82}{0.05} & \tightcell{98.83}{0.03} & \tightcell{98.94}{0.08} & \tightcell{98.96}{0.05} & \tightcell{99.01}{0.04} & \tightcell{\textbf{99.03}}{0} & \tightcell{99.02}{0.04} & \tightcell{98.98}{0.03} & \tightcell{99.03}{0.05} & \tightcell{98.9}{0.13} \\

4.24 / 43.52\% & 20 & \tightcell{98.36}{0.09} & \tightcell{98.38}{0.1} & \tightcell{98.41}{0.1} & \tightcell{98.5}{0.05} & \tightcell{98.53}{0.08} & \tightcell{98.55}{0.12} & \tightcell{98.6}{0.07} & \tightcell{98.78}{0.03} & \tightcell{98.89}{0.03} & \tightcell{98.97}{0.04} & \tightcell{\textbf{99.01}}{0.01} & \tightcell{98.98}{0.03} & \tightcell{98.97}{0.04} & \tightcell{98.97}{0.07} & \tightcell{98.77}{0.03} & \tightcell{98.68}{0.26} \\

1.98 / 20.38\% & 10 & \tightcell{97.42}{0.08} & \tightcell{97.48}{0.08} & \tightcell{97.7}{0.11} & \tightcell{97.96}{0.06} & \tightcell{98.14}{0.09} & \tightcell{98.25}{0.08} & \tightcell{98.21}{0.1} & \tightcell{98.58}{0.03} & \tightcell{98.7}{0.04} & \tightcell{\textbf{98.86}}{0.07} & \tightcell{98.77}{0.05} & \tightcell{98.66}{0.01} & \tightcell{98.67}{0.06} & \tightcell{98.25}{0.12} & \tightcell{96.72}{0.3} & \tightcell{96.19}{0.4} \\

0.94 / 9.69\% & 5 & \tightcell{95.83}{0.34} & \tightcell{96.14}{0.23} & \tightcell{96.71}{0.15} & \tightcell{97.3}{0.12} & \tightcell{97.56}{0.1} & \tightcell{97.79}{0.1} & \tightcell{97.97}{0.15} & \tightcell{98.01}{0.14} & \tightcell{97.94}{0.07} & \tightcell{97.61}{0.16} & \tightcell{\textbf{98.33}}{0.07} & \tightcell{97.85}{0.27} & \tightcell{97.86}{0.17} & \tightcell{97.33}{0.24} & \tightcell{88.98}{1.8} & \tightcell{89.05}{3.08} \\

0.74 / 7.65\% & 4 & \tightcell{95.03}{0.4} & \tightcell{95.51}{0.21} & \tightcell{96.41}{0.3} & \tightcell{97.2}{0.09} & \tightcell{97.47}{0.16} & \tightcell{97.71}{0.14} & \tightcell{97.79}{0.06} & \tightcell{98.14}{0.18} & \tightcell{97.21}{0.24} & \tightcell{96.14}{0.13} & \tightcell{\textbf{98.27}}{0.02} & \tightcell{97.54}{0.25} & \tightcell{97.69}{0.39} & \tightcell{95.36}{0.62} & \tightcell{88.98}{0.82} & \tightcell{85.12}{1.79} \\

0.56 / 5.73\% & 3 & \tightcell{93.9}{0.33} & \tightcell{95.05}{0.35} & \tightcell{95.96}{0.2} & \tightcell{96.82}{0.1} & \tightcell{97.18}{0.13} & \tightcell{97.56}{0.17} & \tightcell{97.66}{0.09} & \tightcell{98.11}{0.11} & \tightcell{97.07}{0.09} & \tightcell{94.31}{1.68} & \tightcell{98.13}{0.14} & \tightcell{\textbf{98.26}}{0.16} & \tightcell{90.91}{2.98} & \tightcell{75.79}{13.32} & \tightcell{74.72}{4.08} & \tightcell{82.99}{1.44} \\

        \hline
    \end{tabular}
    \end{subtable}
    
    \vspace{1em} 
    
    \begin{subtable}{\linewidth}
        \centering
        \caption{Test accuracy with fine-tuning duration $t_{\mathrm{ft}}=3$.}
        \label{tab:carbon_budget-3ft}
        \renewcommand{\arraystretch}{1.2}
        \setlength{\tabcolsep}{4pt}
        \begin{tabular}{|c|c|c|ccccccccccccccc|}
        \hline
        \multirow{2}{*}{\shortstack{\textbf{Carbon} \\ \textbf{Budget ($k$)}}} & \multirow{2}{*}{\shortstack{\textbf{Training} \\ \textbf{Time ($T$)}}} & \multirow{2}{*}{\shortstack{\textbf{No-Slack} \\ \textbf{Baseline}}} & \multicolumn{15}{c|}{\textbf{End time for fine-tuning ($s$)}} \\
        \cline{4-18} 
        & & & \textbf{1} & \textbf{3} & \textbf{5} & \textbf{7} & \textbf{9} & \textbf{10} & \textbf{20} & \textbf{30} & \textbf{40} & \textbf{50} & \textbf{70} & \textbf{90} & \textbf{110} & \textbf{130} & \textbf{150} \\
        \hline
7.98 / 82.0\% & 40 & \tightcell{98.82}{0.06} & \tightcell{97.53}{0.07} & \tightcell{97.69}{0.11} & \tightcell{97.63}{0.3} & \tightcell{97.72}{0.15} & \tightcell{97.78}{0.1} & \tightcell{98.92}{0.03} & \tightcell{98.93}{0.05} & \tightcell{98.98}{0.07} & \tightcell{99}{0.04} & \tightcell{\textbf{99.01}}{0.08} & \tightcell{97.16}{0.19} & \tightcell{97.33}{0.11} & \tightcell{97.97}{0.12} & \tightcell{97.74}{0.55} & \tightcell{97.82}{0.1} \\

6.2 / 63.67\% & 30 & \tightcell{98.66}{0.07} & \tightcell{97.16}{0.12} & \tightcell{97.28}{0.1} & \tightcell{97.48}{0.35} & \tightcell{97.54}{0.21} & \tightcell{97.5}{0.09} & \tightcell{97.36}{0.06} & \tightcell{98.8}{0.02} & \tightcell{98.97}{0.05} & \tightcell{99.02}{0.02} & \tightcell{99.02}{0.04} & \tightcell{99.05}{0.01} & \tightcell{\textbf{99.09}}{0.05} & \tightcell{99.06}{0.06} & \tightcell{99.05}{0.07} & \tightcell{99.01}{0.04}\\

4.24 / 43.52\% & 20 & \tightcell{98.36}{0.09} & \tightcell{96.47}{0.18} & \tightcell{96.72}{0.14} & \tightcell{96.74}{0.04} & \tightcell{96.77}{0.17} & \tightcell{96.8}{0.39} & \tightcell{97.03}{0.14} & \tightcell{97.35}{0.21} & \tightcell{98.9}{0.04} & \tightcell{98.89}{0.05} & \tightcell{\textbf{98.99}}{0.01} & \tightcell{98.9}{0.01} & \tightcell{94.22}{1.18} & \tightcell{93.21}{2.6} & \tightcell{94.97}{1.54} & \tightcell{92.41}{1.34} \\

1.98 / 20.38\% & 10 & \tightcell{97.42}{0.08} & \tightcell{93.5}{0.83} & \tightcell{94.05}{0.37} & \tightcell{95.18}{0.49} & \tightcell{95.65}{0.31} & \tightcell{95.77}{0.22} & \tightcell{96.13}{0.42} & \tightcell{97.1}{0.31} & \tightcell{97.45}{0.27} & \tightcell{\textbf{98.82}}{0.04} & \tightcell{98.71}{0.04} & \tightcell{98.57}{0.16} & \tightcell{98.1}{0.24} & \tightcell{98.06}{0.3} & \tightcell{95.8}{1.59} & \tightcell{92.24}{1.4}\\

1.77 / 18.23\% & 9 & \tightcell{97.37}{0.08} & \tightcell{91.44}{1.64} & \tightcell{91.84}{1.18} & \tightcell{91.32}{0.96} & \tightcell{91.52}{0.55} & \tightcell{89.2}{1.34} & \tightcell{87.31}{1.8} & \tightcell{92.07}{1.52} & \tightcell{94.2}{0.43} & \tightcell{\textbf{98.63}}{0.07} & \tightcell{98.52}{0.17} & \tightcell{98.38}{0.04} & \tightcell{97.78}{0.31} & \tightcell{97.76}{0.28} & \tightcell{92.93}{0.82} & \tightcell{92.41}{1.71}\\

0.94 / 9.69\% & 5 & \tightcell{95.83}{0.34} & \tightcell{82.04}{0.98} & \tightcell{84.91}{1.16} & \tightcell{88.55}{0.48} & \tightcell{84.64}{0.85} & \tightcell{91.32}{0.96} & \tightcell{90.05}{0.9} & \tightcell{92.12}{0.12} & \tightcell{94.21}{0.46} & \tightcell{88.91}{1.6} & \tightcell{\textbf{97.13}}{0.15} & \tightcell{22.7}{15.46} & --- & --- & --- & --- \\

0.74 / 7.65\% & 4 & \tightcell{95.03}{0.4} & \tightcell{78.99}{0.94} & \tightcell{82.87}{1.56} & \tightcell{95.14}{0.47} & \tightcell{90.88}{0.71} & \tightcell{97.39}{0.14} & \tightcell{91.09}{0.77} & \tightcell{95.7}{0.25} & \tightcell{96.64}{0.28} & \tightcell{\textbf{97.89}}{0.09} & \tightcell{96.51}{0.36} & --- & --- & --- & --- & --- \\
        \hline
    \end{tabular}
    \end{subtable}
\end{table*}

\section{Carbon-Aware FL Scheduler and Algorithm}
\label{sec:algo}

\begin{figure}[t]
    \centering
    \includegraphics[width=0.75\columnwidth]{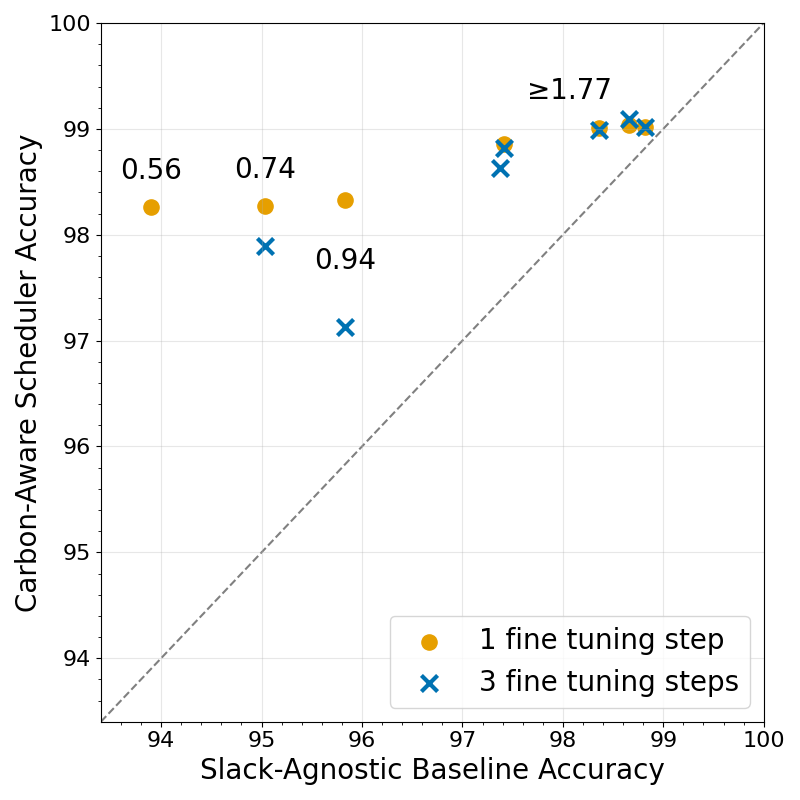}
    \caption{Final test accuracy for our carbon-aware scheduler (y-axis) and the slack-agnostic baseline (x-axis) across varying carbon budgets. Each point corresponds to the best end time for fine-tuning configuration (from Table~\ref{tab:combined_table}) for a given budget level and is annotated with its associated budget.}
    \label{fig:50_0.5_slack}
\end{figure}

Building on previous sections, we now propose a unified \emph{carbon-aware FL scheduling} formulation that integrates:\\
(i) \emph{slack time} $t_\mathrm{sl} \geq 0$, extending training to $T + t_\mathrm{sl}$ rounds;\\
(ii) \emph{$\alpha$-fair carbon allocation}, parameterized by $\alpha \in (0, 1]$;\\
(iii) \emph{fine-tuning} of duration $t_{\mathrm{ft}}$, which selects all clients.

Since fine-tuning may also consume a significant share of the carbon budget, we optimize its temporal placement. Formally, we define a movable fine-tuning window $\mathcal{F}(s) = \{ T+s-t_{\mathrm{ft}}+1, \dots, T+s \}$ and treat its end time $s \in \{1, \dots, t_\mathrm{sl}\}$ as a decision variable. We then jointly optimize the scheduling matrix $A \coloneqq (a_{c}^{(t)})_{c,t}$ and the fine-tuning placement $s$ under a global carbon budget constraint:
\begin{maxi!}
    {A, s}{\sum_{c=1}^{K} \biggl( \sum_{t=1}^{T + t_\mathrm{sl}} (g_{\max} - g_{c}^{(t)}) a_{c}^{(t)} \biggr)^{\alpha} \label{opt:tuning:objective}}
    {\label{opt:tuning}}{}
    \addConstraint{\sum_{c=1}^{K} \sum_{t=1}^{T + t_\mathrm{sl}} g_{c}^{(t)} a_{c}^{(t)}\leq k \label{opt:constr:budget}}{}
    \addConstraint{\mathcal{F}(s)= \{T+s-t_{\mathrm{ft}}+1, \dots, T+s\} \label{opt:constr:ft_window}}{}
    \addConstraint{a_{c}^{(t)}\in \{0,1\}, \mspace{5mu} \forall c,t < T+s-t_{\mathrm{ft}}+1 \label{opt:constr:pre_ft}}{}
    \addConstraint{a_{c}^{(t)}=1, \mspace{97mu}\quad \forall c,t \in \mathcal{F}(s) \label{opt:constr:alloc_ft}}{}
    \addConstraint{a_{c}^{(t)}=0, \mspace{89mu}\quad \forall c,t > T + s \label{opt:constr:end_train}}{}.
\end{maxi!}
The fine-tuning phase, during which all clients are selected (constraint~\eqref{opt:constr:alloc_ft}), is included in the global carbon budget $k$ (constraint~\eqref{opt:constr:budget}), and training concludes at the last fine-tuning round $T+s$ (constraint~\eqref{opt:constr:end_train}).

Problem~\eqref{opt:tuning} is a non-convex integer nonlinear program, formulated as a maximization problem, and is NP-hard by reduction to the 0–1 knapsack problem.
At the scale considered in our experiments, it can be solved to numerical optimality using the MOSEK solver through the CVXPY library, in under a minute.
For larger-scale instances, the monotone and submodular structure of the objective under a single-knapsack constraint admits a polynomial-time greedy algorithm, 
which iteratively selects client–time slot pairs with the highest marginal carbon utility per unit cost and achieves a tight $(1-1/e)$ approximation ratio~\cite{Krause_Golovin_2014}.

Algorithm~\ref{alg:general_algo} summarizes our \emph{carbon-aware FL training} procedure. The server leverages our carbon-aware scheduler (Problem~\eqref{opt:tuning}) to precompute the client–time slot allocation (line~\ref{alg-line:scheduler}). Only the selected clients ($a_{c}^{(t)}=1$) participate in training and send their model updates to the server (lines~\ref{alg-line:client_opt}--\ref{alg-line:client_update}).  
To correct for selection bias, the server applies U-FedAvg aggregation 
(lines~\ref{alg-line:agg_ufedavg}--\ref{alg-line:server_opt_ufedavg}), switching to standard FedAvg during fine-tuning when all clients are selected (lines~\ref{alg-line:agg_fedavg}--\ref{alg-line:server_opt_fedavg}).

\section{Model Accuracy under Carbon Constraints}
\label{sec:exp_real_carbon_data}
 
We benchmark our carbon-aware scheduler against a slack-agnostic baseline that runs standard FedAvg and selects all clients in every round until the carbon budget~$k$ is exhausted. This baseline achieves highest statistical representativeness, fastest per-round convergence (as full-client gradients provide the steepest unbiased descent direction), and uniform client selection ($\rho_{\text{H}}=0$). It is a provably optimal baseline under stationary carbon intensity or sufficiently large budgets.

Figure~\ref{fig:50_0.5_slack} and Table~\ref{tab:combined_table} report the final test accuracy of our carbon-aware algorithm under a wide range of carbon budgets and fine-tuning end times. Budgets vary from low-carbon regimes (5\% of the full-budget reference) to high-carbon regimes (up to 82\%). For each budget, we vary the fine-tuning end time $s \in \{ 1, 3, \dots, 150 \}$ and consider two fine-tuning durations $t_{\mathrm{ft}}\in \{ 1, 3 \}$. We fix the fairness parameter to $\alpha = 0.1$, identified in Figure~\ref{fig:alphaF_var} as providing an effective trade-off between test accuracy and carbon efficiency.

Figure~\ref{fig:50_0.5_slack} compares the final test accuracy of our carbon-aware scheduler (y-axis) against the slack-agnostic baseline (x-axis) across matching carbon budgets. Each point is annotated with its associated budget level. 
Across most regimes, our scheduler consistently outperforms the baseline, particularly under tight budget constraints: at 5.73\% and 7.65\% of the full-budget reference, our scheduler achieves gains of +4.36 and +3.24 pp, respectively. As the budget increases, the margin narrows (e.g., +0.2 pp at 82\%), showing that the baseline becomes strongly competitive in high-budget settings.

Table~\ref{tab:combined_table} analyzes the impact of fine-tuning end time~$s$, fine-tuning duration~$t_{\mathrm{ft}}$, and slack time~$t_\mathrm{sl}$ 
on final test accuracy. 
At low carbon budgets (e.g., 9.69\%), shorter training durations ($T+s \leq 55$) with moderate slack ($t_\mathrm{sl} \leq 50$) achieve higher accuracy (+2.5 pp over no slack), as longer durations lead to sparser client selection and increase the risk of forgetting initial slots.
Similarly, at low carbon budgets (9.69\%), a shorter fine-tuning duration ($t_{\mathrm{ft}}=1$) achieves +1.2 pp higher accuracy than $t_{\mathrm{ft}}=3$, likely due to the additional carbon cost of longer fine-tuning durations.
In contrast, at high carbon budgets (e.g., $63.67\%$), longer training durations ($T + s \geq 100$) with larger slack ($t_\mathrm{sl} \geq 70$) yield higher accuracy (+0.37 pp over no slack), as the extended horizon provides more opportunities to allocate low-carbon, high-utility training slots.

\section{Conclusion}
\label{sec:conclusion}

This paper investigated emissions reduction in federated learning through carbon-aware training scheduling. Beyond quantifying potential carbon savings, we highlighted key challenges in balancing environmental objectives with learning performance, including statistical heterogeneity, selection bias, and temporal imbalance of training slots. By introducing a scheduler that integrates slack time, fairness-aware allocation, and fine-tuning, we take a first step toward addressing these challenges. We believe that exploring the trade-off between emission reductions and learning performance is key to developing sustainable, carbon-efficient federated learning systems.

\bibliographystyle{IEEEtran}
\bibliography{conference_101719}

\end{document}